\documentclass[journal]{IEEEtran}
\usepackage{cite}
\usepackage{graphicx}
\usepackage{amsmath,amsfonts,amssymb}
\usepackage{algorithmic}
\usepackage{textcomp}
\usepackage{xcolor}
\usepackage{hyperref}

\title{Continual Learning-Based Unified Model for Unpaired Image Restoration Tasks}

\author{
\IEEEauthorblockN{Kotha Kartheek, Lingamaneni Gnanesh Chowdary, Snehasis Mukherjee, \IEEEmembership{Senior Member, IEEE}} \\
\IEEEauthorblockA{Shiv Nadar Institution of Eminence, Delhi NCR, India\\
Email: \{kk746, lc607, snehasis.mukherjee\}@snu.edu.in}
}
\begin{document}
\maketitle

\begin{abstract}
Restoration of images contaminated by different adverse weather conditions such as fog, snow, and rain is a challenging task due to the varying nature of the weather conditions. Most of the existing methods focus on any one particular weather conditions. However, for applications such as autonomous driving, a unified model is necessary to perform restoration of corrupted images due to different weather conditions. We propose a continual learning approach to propose a unified framework for image restoration. The proposed framework integrates three key innovations: (1) Selective Kernel Fusion layers that dynamically combine global and local features for robust adaptive feature selection; (2) Elastic Weight Consolidation (EWC) to enable continual learning and mitigate catastrophic forgetting across multiple restoration tasks; and (3) a novel Cycle-Contrastive Loss that enhances feature discrimination while preserving semantic consistency during domain translation. Further, we propose an unpaired image restoration approach to reduce the dependance of the proposed approach on the training data. Extensive experiments on standard benchmark datasets for dehazing, desnowing and deraining tasks demonstrate significant improvements in PSNR, SSIM, and perceptual quality over the state-of-the-art. The code will be available after acceptance.
\end{abstract}

\begin{IEEEkeywords}
Image restoration, cycleGAN, Selective Kernel Networks, Continual Learning, Contrastive Learning.
\end{IEEEkeywords}

\section{Introduction}
\IEEEPARstart{O}{utdoor} vision systems often face significant challenges when capturing images in adverse weather conditions such as haze/ fog, rain, and snow. These conditions severely degrade visibility and contrast, hampering crucial applications, including traffic monitoring, object detection, autonomous driving, and outdoor surveillance. Image restoration under such adverse conditions remains an ill-posed problem that requires sophisticated computational approaches to recover clean, visually appealing images.

Traditional restoration approaches primarily rely on hand-crafted priors, such as the Dark Channel Prior (DCP) \cite{darkchannelprior1, ref_ffa_net}, Color Attenuation Prior (CAP) \cite{cap}, and Non-Local Color Prior (NCP) \cite{ncp}. While these physical model-based methods can improve visibility to some extent, their dependency on specific priors often restricts their ability to generalize across diverse real-world scenarios. Additionally, these methods typically require meticulous parameter tuning, limiting their practicality for automated systems.

Recent advancements in deep learning have revolutionized image restoration techniques \cite{survey}. Convolutional Neural Networks (CNNs) \cite{cnn, cnn1, cnn2, cnn3} and Generative Adversarial Networks (GANs) \cite{GAN2, aglc_gan, jaiSurya_0, CHAITANYA2021103014, cclgan} have demonstrated remarkable capabilities in addressing weather-related degradations. Models such as RESCAN \cite{RE-SCAN} effectively learn hierarchical features for image restoration. However, most supervised learning methods require paired datasets (degraded-clean image pairs), which are challenging to get in real-world settings.

To overcome these limitations, unsupervised (unpaired) and semi-supervised learning paradigms have emerged as promising alternatives. CycleGAN-based approaches \cite{aglc_gan, CycleGAN_deHAze, cclgan} attempt to bridge the gap between source and target domains through cycle consistency, enabling unpaired image-to-image translation. However, conventional CycleGAN implementations often struggle with preserving fine details and maintaining global consistency, resulting in suboptimal restoration quality.

GANs with attention mechanisms have shown promising results for single-domain tasks, with models such as AGLC-GAN demonstrating remarkable performance in image dehazing in unpaired setup \cite{aglc_gan}. However, AGLC-GAN typically lacks the flexibility to handle multiple weather conditions simultaneously. Furthermore, when adapting such models to new weather conditions, catastrophic forgetting - where a model loses previously learned knowledge - becomes a significant challenge.

Motivated by AGLC-GAN, we propose an unpaired cycleGAN-based model for image restoration. Further, we address multiple weather-related distortions in images through a continual learning framework. Moreover, we integrate contrastive learning \cite{contrastivelearning1, contrastivelearning2, cclgan} to enhance discriminative feature learning across different weather conditions and incorporate Elastic Weight Consolidation (EWC) \cite{continualLearning, continualLearning1, continualLearning2} to mitigate catastrophic forgetting when sequentially learning new restoration tasks.

The main contributions of this work are:
\begin{itemize}
    \item Proposed \textbf{DA-AGLC-GAN} model, a novel unified network for adverse weather removal that integrates SK Fusion Layers and cycle-contrastive learning to improve restoration quality across different weather conditions.
    \item Proposed a continual learning framework utilizing EWC to enable the model to sequentially learn dehazing, deraining, and desnowing while effectively mitigating catastrophic forgetting.
    \item Comprehensive experimental evaluations on benchmark datasets for dehazing, desnowing, and deraining, demonstrating the superior performance of our proposed method compared to state-of-the-art approaches, both for individual tasks and in a continual learning setting.
    \item Detailed ablation studies quantifying the contribution of each proposed component to the overall performance of our framework.
\end{itemize}


\section{Background and Related Work}
\label{sec:related_work}
Adverse weather removal has been a problem extensively studied during the last few years, to mitigate visual distortions caused by atmospheric interference \cite{survey}. Early approaches predominantly relied on physical priors, such as the Dark Channel Prior (DCP) \cite{ref_ffa_net,darkchannelprior1,ffa_net} for dehazing and low-rank representations for rain streak removal, to eliminate degradations. Consequently, deep learning-based solutions have emerged as the dominant paradigm, such as Convolutional Neural Networks (CNNs) \cite{cnn,cnn1,cnn2,cnn3} and attention mechanisms to develop architectures such as DehazeNet \cite{DeHazeNet,DeHazeNet1} and RESCAN \cite{RE-SCAN}, which effectively learn hierarchical features for image restoration. More recent advancements have integrated Generative Adversarial Networks (GANs) \cite{GAN1} to capture global dependencies and enhance restoration fidelity. Despite significant progress, challenges remain in handling highly non-uniform degradations, ensuring robust generalization across domains, and achieving computational efficiency for real-time applications in autonomous systems and surveillance.

\subsection{Generative Adversarial Networks (GANs)}
Early works of GAN based image restoration, such as Wasserstein GAN \cite{2017wassersteingan} and ESRGAN \cite{GAN2} improved training stability and perceptual quality, respectively. In the domain of adverse weather restoration, GAN-based methods have been successfully applied to single image dehazing \cite{cycleGAN, aglc_gan}. The Omni-kernel Network \cite{Cui_Ren_Knoll_2024} have demonstrated promising results in real-time image restoration while maintaining efficiency. Despite these advances, most GAN-based approaches struggle with domain adaptation and require paired data, which is difficult to acquire in real-world scenarios.

\subsection{Unpaired Adverse Weather Removal}
The lack of paired real-world data for adverse weather conditions has driven the development of unsupervised learning techniques. These methods utilize adversarial training and domain adaptation. Recent work such as UCL-Dehaze \cite{UCL-Dehaze} has effectively integrated domain adaptation with self-supervised strategies to improve feature learning without the need for paired data. The work on Disentangled Bad Weather Removal GAN for Pedestrian Detection \cite{disentangled_bad_weather} extracts weather-specific layers for unified removal of rain and haze, thereby enhancing pedestrian detection performance using Cycle-GAN.

AGLC-GAN \cite{aglc_gan} builds on the CycleGAN framework by introducing a global-local attention mechanism for more effective haze removal. However, AGLC-GAN fails to preserve fine structural details and maintain color accuracy, resulting in suboptimal restoration, especially for deraining and desnowing. In \cite{FAN2025150}, Fan et al. introduced the dual-branch collaborative unpaired dehazing model to alleviate the domain deviation problem. Beyond adversarial learning, contrastive learning \cite{contrastivelearning1} and pseudo-labeling \cite{pseudo-labeling} have shown promise in further improving adverse weather removal. By exploiting relationships between different degraded and clean image distributions, these approaches enable models to learn more discriminative representations without explicit supervision.

\subsection{Contrastive Learning}
Contrastive learning \cite{UCL-Dehaze,contrastivelearning1,contrastivelearning2} has gained significant attention in computer vision by enabling models to learn feature representations without requiring labeled data. Although widely used in high-level vision tasks, recent studies have explored its application in low-level image restoration, such as dehazing, deraining, and super-resolution. Contrastive learning leverages contrastive loss by treating clean images as positive samples and degraded images as negatives, guiding the network to identify their dissimilarity. More recent work has extended this paradigm to unpaired settings \cite{contrastivelearning2}, utilizing reconstructed images or natural scene variations as contrastive pairs to improve generalization. 

Recent studies, such as Cycle-Contrastive Adversarial Learning \cite{cclgan} and Model Contrastive Learning for Image Restoration (MCLIR) \cite{contrastivelearning3}, have successfully integrated contrastive objectives with GAN frameworks to boost performance across dehazing, deraining, and super-resolution tasks. MCLIR introduces ``learning from history", where past latent models serve as dynamic negative samples, refining feature representations more effectively than traditional approaches. These approaches enable the network to learn shared, discriminative features across diverse degradations, while requiring careful balancing to avoid over-smoothing and loss of fine details. Furthermore, most existing contrastive learning frameworks for image restoration focus on single-domain tasks, lacking mechanisms to transfer knowledge across different weather conditions effectively.

\subsection{Continual Learning}
\label{subsec:continual_learning}
Continual learning (CL) enables models to sequentially learn multiple tasks while retaining previously acquired knowledge, which is critical for achieving robust restoration of images corrupted by adverse weather conditions. A major challenge in CL is \textbf{catastrophic forgetting}, where adapting to new tasks leads to a decline in performance on earlier tasks. To address this, various strategies have been proposed, including memory-based methods that store and replay past experiences~\cite{continualLearning1, replayMethod_CL_2} and regularization-based approaches such as Elastic Weight Consolidation (EWC)~\cite{continualLearning,continualLearning1,continualLearning2}. For learning dehazing, desnowing and deraining tasks, EWC offers an efficient solution by constraining parameter updates to preserve critical knowledge without the overhead of memory storage \cite{cheng2024continualallinoneadverseweather}. In Cheng et al. \cite{cheng2024continualallinoneadverseweather}, a framework utilizing knowledge replay within a unified network was proposed; however, this approach required paired data and replay mechanisms. Recent studies \cite{continualLearning3} have demonstrated the effectiveness of regularization-based methods, such as EWC, in GAN-based image generation, highlighting their capacity to maintain feature consistency across tasks. By incorporating EWC into our framework, we not only enhance generalization but also mitigate catastrophic forgetting, enabling the model to seamlessly adapt to new restoration tasks without sacrificing performance on previously learned tasks.

\section{Proposed Methodology}
\label{sec:proposed_methodology}
We begin with the Attention-based global-local cycle-consistent generative adversarial networks (AGLC-GAN) \cite{aglc_gan} which was used for \textbf{Single-Image deHazing} as our base model and systematically enhance its architecture to improve performance across multiple weather conditions. We refine the network structure by incorporating \textbf{Selective Kernel (SK) Fusion Layers} into AGLC-GAN, as explained in Subsection \ref{subsec:selective_kernel}, to enhance the model’s ability to leverage multi-scale information. Additionally, we introduce a \textbf{Feature Refinement Head}, utilizing cycle-contrastive learning, as described in Subsection \ref{subsec:ccl_loss}, to improve the model’s capacity to maintain better alignment with the semantic content and structure of input images.

\subsection{Backbone - AGLC-GAN Model}
\label{subsec:aglc_gan}
The AGLC-GAN architecture leverages a CycleGAN framework with the following components: (1) A CycleGAN architecture with an autoencoder-like generator, (2) Improved Feature Attention Mechanism combining channel and pixel attention, (3) Global–Local discriminator to handle spatially varying features, (4) Dynamic feature enhancement block, and (5) adaptive mixup module. The system consists of two generators, \(G_A\) and \(G_B\), and two discriminators, \(D_A\) and \(D_B\). \(G_A\) transforms hazy images into the clear domain, while \(G_B\) transforms clear images back into the hazy domain. The process ensures consistency using both \textit{Cycle Consistency Loss} and \textit{Cyclic Perceptual Consistency Loss}.
\begin{figure}
    \centering
    \includegraphics[width=0.95\columnwidth]{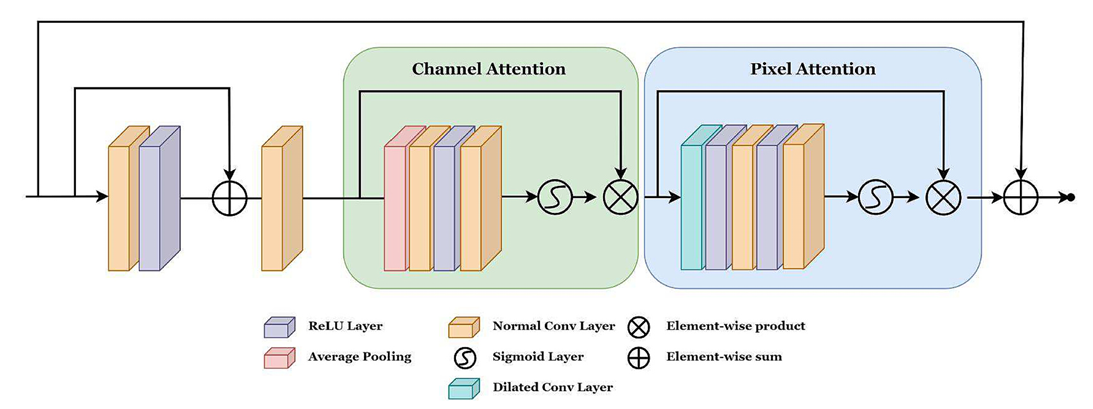}
    \caption{Architecture of the proposed Feature Attention Module.}
    \label{fig:fa_arc}
\end{figure}

\subsubsection{Generator Architecture}
The generator consists of three modules: an encoder, a feature transformation module, and a decoder. The encoder downsamples the input using a 4× downsampling operation to extract latent features. Each downsampling block includes a convolution layer (kernel size = \(3 \times 3\), stride = 2), followed by Instance Normalization and a ReLU activation layer. A larger kernel size (\(7 \times 7\)) is used in the initial convolution to capture broader contextual information.

The Feature Transformation Module consists of three key components: (i) Channel Attention (CA), (ii) Pixel Attention (PA), and (iii) Dynamic Feature Enhancement (DFE). The Feature Attention (FA) block integrates CA and PA mechanisms to selectively emphasize relevant features inspired from FFA-NET \cite{ref_ffa_net}, while the DFE module employs Deformable Convolutional Networks V2 (DCNv2) \cite{acerNet} to capture structural details and handle geometric transformations. The generator is designed with six FA blocks and two DFE modules, ensuring effective feature extraction and transformation.

The CA mechanism assigns importance to each channel based on its contribution to the distortion level in the image. The attention weights are computed as follows:
\begin{equation}
CA = \sigma (Conv(\delta(Conv(g_c)))),
\end{equation}
where \( g_c \) is the globally pooled feature, \( \delta \) represents the ReLU activation function, and \( \sigma \) is the sigmoid activation function. The refined feature map \(F^*_c\) is then obtained by element-wise multiplication with the original feature \( F_c\):
\begin{equation}
F^*_c = \text{CA} \otimes F_c.
\end{equation}

The PA mechanism focuses on highly distorted regions using dilated convolutions inspired by \cite{dilated_net}. The attention weights are computed as:
\begin{equation}
    PA = \sigma (Conv(\delta(Conv(\delta(DilatedConv(F^*)))))),
\end{equation}
where \( F^* \) is the input feature map, and the PA weights are applied element-wise:
\begin{equation}
\tilde{F} = F^* \otimes PA.
\end{equation}

The DFE module utilizes DCNv2 \cite{DCN_v2} to refine structural details by dynamically adjusting receptive fields. It consists of two deformable convolution layers, each with learned spatial offsets and modulation masks. The deformable convolution operation is formulated as:  
\begin{equation}
    y(p) = \sum_{k=1}^{K} w_k \cdot x(p + p_k + \Delta p_k) \cdot \Delta m_k,
\end{equation}
where $x(p)$ and $y(p)$ denote the features at location $p$ from the input feature maps x and output feature maps $y$, respectively and \( w_k \) represents kernel weights, \( p_k \) is the default offset, \( \Delta p_k \) and \( \Delta m_k \) are learnable parameters  for the k-th location, and \( x \) and \( y \) represent the input and output features.

In our implementation, two deformable convolution layers (\texttt{dcn1} and \texttt{dcn2}) are employed with a kernel size of \( k \times k \). The corresponding offset and mask convolutions are defined for each deformable layer, where the offset convolutions predict \( 2k^2 \) displacement values, and the mask convolutions generate \( k^2 \) modulation values. These components collectively enable the network to adaptively refine features by focusing on structurally significant regions of the input. The overall architecture of the proposed Feature Attention Module is shown in Figure \ref{fig:fa_arc}. 

The decoder upsamples latent features back to the target domain using 4× upsampling. It consists of two upsampling modules with deconvolution layers ($3 \times 3$ kernel, stride 2), each followed by Instance Normalization and a ReLU activation function. Finally, a convolution layer ($7 \times 7$ kernel, stride 1) is applied, followed by a Tanh activation to produce the final output.

\subsubsection{Discriminator}
The Discriminator consists of two parts: the Global Discriminator and the Local Discriminator, both having the same architecture. The Global Discriminator processes the entire image to capture the global characteristics of the image, while the Local Discriminator processes randomly cut patches from the image to capture local features and variations.
\begin{figure}
    \centering
    \includegraphics[width=0.95\columnwidth]{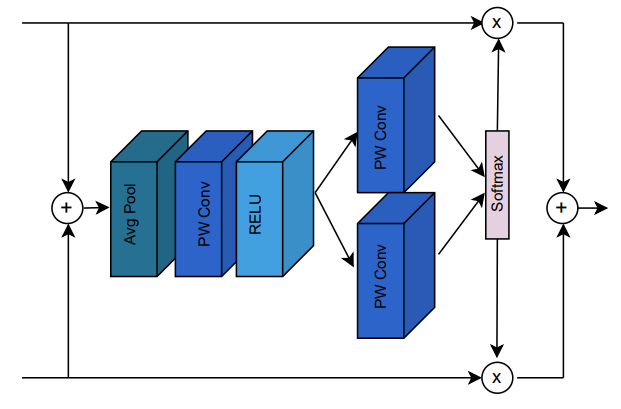}
    \caption{Architecture of the SK Fusion module.}
    \vspace{0.75em}
    \label{fig:skfusion}
\end{figure}

The Global Discriminator takes the whole image as input. In contrast, the Local Discriminator takes 3 randomly cut patches of size $64 \times 64$ from the image. We use 3 patches to balance processing time with sufficient variation in the local features.

\textbf{Global Discriminator} operates on the entire image, passing it through 5 convolutional blocks with Instance Normalization and Leaky ReLU, reducing the input image size ($3\times H \times W$) to a downsampled feature map ($1\times H^* \times W^*$).

\textbf{Local Discriminator} operates on the randomly selected $64 \times 64$ patches providing detailed information about the local features of the image.

The outputs from both the Global and Local Discriminators are flattened and concatenated into a single vector, which is then passed through a sigmoid activation function. The resulting vector is used to classify whether the input image is real or fake.

\subsection{Selective Kernel Fusion Layers}
\label{subsec:selective_kernel}
Selective Kernels (SKs), introduced in \cite{sk_ref}, dynamically adjust receptive fields by integrating multi-scale feature representations. Inspired by the human visual system, SKs utilize multiple kernel sizes to extract features and employ an attention-driven selection mechanism to prioritize the most relevant spatial information. This adaptability allows the network to capture both fine-grained local details and broader contextual structures, enhancing its effectiveness in image restoration tasks.

Adverse weather conditions such as haze, rain, and snow introduce distortions with varying scales and intensities, making it essential to design an architecture capable of adaptive feature fusion. Instead of conventional mix-up modules, we integrate \textbf{SK Fusion Layers}, motivated by DRGNet \cite{skFusion} to enhance the structural composition of the AGLC-GAN, enabling the model to process multi-scale information more effectively. These layers dynamically merge feature maps, ensuring that spatial and contextual details are preserved and enhanced for improved image restoration.  

The proposed SK Fusion module adaptively refines feature representations by selectively combining information from skip connections and the main network path. Given input feature maps \( x_1 \) (skip connection) and \( x_2 \) (main path), the fusion process follows:

\textbf{Feature Combination}: The input feature maps \( x_1 \) and \( x_2 \) are concatenated, and globally processed by reshaping.  

\textbf{Global Attention Mechanism}: A Global Average Pooling (GAP) operation extracts spatial statistics, which are then passed through a lightweight multi-layer perceptron (MLP) with 2 Convolution layers and 1 ReLU layer. The generated attention weights \(\{a_1, a_2\}\) are computed as:  
\begin{equation}
    \{a_1, a_2\} = \text{Softmax} \left( \text{MLP} \left( \text{GAP}(x_1 + x_2) \right) \right).
\end{equation}

\textbf{Weighted Fusion}: The final output is obtained through a weighted sum of the input features, and the generated attention weights:
\begin{equation}
    y = a_1 x_1 + a_2 x_2.
\end{equation}  

We incorporate SK Fusion layers at two key locations in the generator architecture. The first SK Fusion layer is used between the third (last) downsampling layer and the first upscaling layer, where it merges the deepest encoder features with the DFE module output. This fusion ensures that both extracted and refined structural details are effectively integrated, enhancing the network’s ability to recover fine textures. The second SK Fusion layer is introduced between the second (pre-ultimate) downsampling layer and the second upscaling layer. This placement strengthens multi-scale feature representations, enabling the model to reconstruct images with improved preservation of details, and allows the model to selectively emphasize the most informative feature scales, significantly improving its adaptability to different weather-based distortions and enhancing overall image restoration performance. The architecture of SK Fusion Module is shown in Fig \ref{fig:skfusion}.
\begin{figure*}
    \centering
    \includegraphics[width=\textwidth]{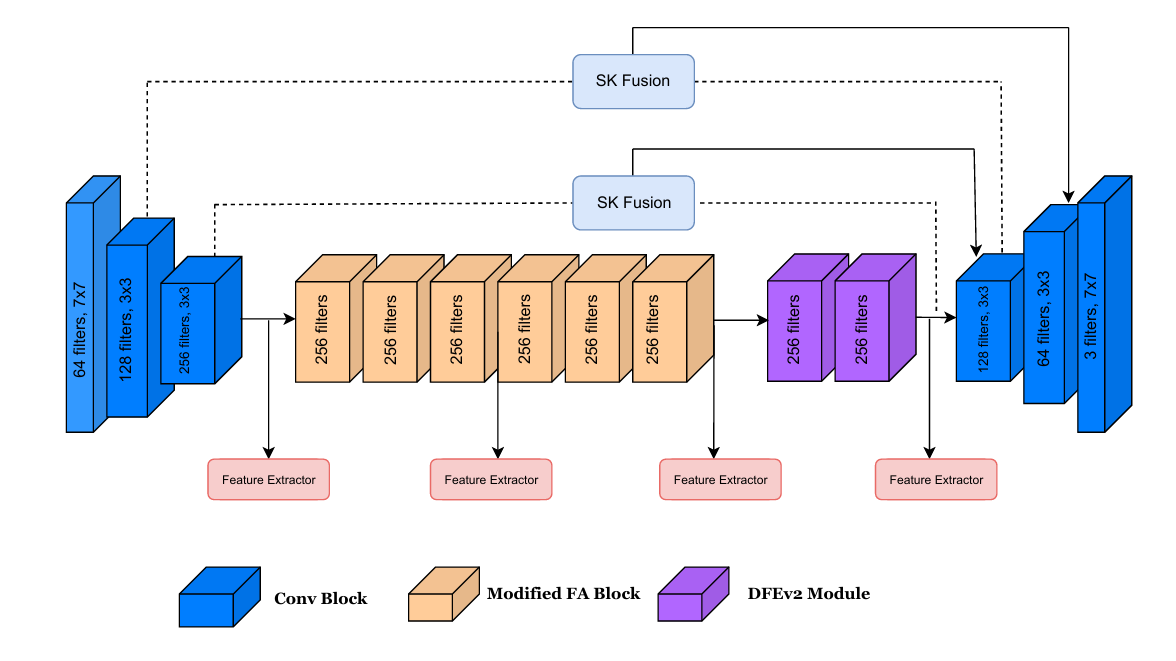}
    \caption{Overall Architecture of the proposed Generator Model.}
    \label{fig:new_arc}
\end{figure*}

\subsection{Cycle-Contrastive Loss}
\label{subsec:ccl_loss}
Contrastive learning is a technique that focuses on extracting meaningful representations by distinguishing between similar (positive) and dissimilar (negative) instance pairs. For our model, contrastive learning enforces content preservation while refining structural details during tasks such as dehazing and deraining. Specifically, we incorporate \textbf{Cycle-Contrastive Loss (CCL)}, motivated by CCLGAN \cite{cclgan} to encourage feature alignment between the input and output images, ensuring that generated images retain semantic integrity and spatial consistency.

The CCL enhances model's ability to maintain content consistency during the cycle of image transformations, such as converting a rainy image (\( r \)) into a rain-free image (\( n \)). The key objective of CCL is to align the generated output with the corresponding content features from the input image while maximizing mutual information between the input and output at the same spatial locations.

To achieve this, content features are extracted from both input image and generated image using the encoder of the generators \textbf{ \(G_A\)} and \textbf{\(G_B\)}, respectively. Feature maps are extracted from predefined layers (\texttt{nce\_layers}) and reshaped into 64 spatial patches. These patches are further transformed into a new feature space using MLP, and normalized using L2 normalization, ensuring unit-length embeddings. The \textbf{positive samples} are the samples corresponding to the same spatial locations in the input and output images, while \textbf{negative samples} are chosen from unrelated locations.

The contrastive loss is then computed using cosine similarity between the query feature \( q \) (from the output image) and key features \( k^+ \) (positive) and \( k^- \) (negative). The similarity is controlled by a temperature parameter \( \tau \), which scales the loss. The Cycle-Contrastive Loss is defined as:
\begin{equation}
\label{lccl}
L_{\text{CCL}} = - \log \left[ \frac{\exp\left(\frac{\text{Sim}(q, k^+)}{\tau}\right)}{\exp\left(\frac{\text{Sim}(q, k^+)}{\tau}\right) + \sum_{i=1}^{N} \exp\left(\frac{\text{Sim}(q, k^-_i)}{\tau}\right)} \right],
\end{equation}
where $Sim(u, v) = \frac{u^\top v}{\|u\|\|v\|}$ denotes the cosine similarity, \( \tau \) is the temperature parameter, set to \( 0.07 \) as used in \cite{cclgan}, \( N \) is the number of negative samples (64).

The Feature extractors are placed in 4 different places across the generator from which the content features are extracted. The first one is placed after downsampling layers, second in between FA modules, third after the FA modules and the final one before the upsampling layers. Our overall model architecture Discriminative Analysis AGLC-GAN (\textbf{DA-AGLC-GAN}) integrating CCL extraction modules and SK Fusion to AGLC-GAN is depicted in Fig. \ref{fig:new_arc}.

\subsection{Overall Loss}
\label{over_loss}
The loss function for the proposed model comprises 4 parts: (1) Adversarial loss and (2) Cycle Consistency loss from the original CycleGAN Model, (3) Cyclic Perceptual Consistency loss from Cycle-Dehaze \cite{pix2pix} to achieve perceptual similarity between the generated and the original images and (4) Cycle Contrastive Loss.
\subsubsection{Adversarial Loss}
For the generators \(G_A\) (hazy-to-clear) and \(G_B\) (clear-to-hazy) and their corresponding discriminators \(D_B\) and \(D_A\), the adversarial loss is formulated as the least-square loss. Specifically, the loss for \(G_A\) with discriminator \(D_B\) is,
\begin{align}
\mathcal{L}_{\text{GAN}}(G_A, D_B, A, B) = \mathbb{E}_{B \sim p_{\text{data}}(B)} \left[ \log D_B(B) \right]  \\
+ \mathbb{E}_{A \sim p_{\text{data}}(A)}  \left[ \log \left( 1 - D_B(G_A(A)) \right) \right].
\end{align}

Similarly, for \(G_B\) and \(D_A\), the adversarial loss is:
\begin{align}
\mathcal{L}_{\text{GAN}}(G_B, D_A, B, A) = \mathbb{E}_{A \sim p_{\text{data}}(A)} \left[ \log D_A(A) \right] \\
+ \mathbb{E}_{B \sim p_{\text{data}}(B)} \left[ \log \left( 1 - D_A(G_B(B)) \right) \right].
\end{align}
The total adversarial loss is the sum of these two components:
\begin{align}
    \mathcal{L}_{\text{GAN}} = \mathcal{L}_{\text{GAN}}(G_A, D_B, A, B) + \mathcal{L}_{\text{GAN}}(G_B, D_A, B, A).
\end{align}

\subsubsection{Cycle Consistency Loss}
To ensure consistency between the input domain and the target domain, we use the Cycle Consistency loss, formulated as:
\begin{multline}
\mathcal{L}_{\text{cyc}}(G_A, G_B) = \mathbb{E}_{A \sim p_{\text{data}}(A)} \left[ \| G_B(G_A(A)) - A \|_1 \right] \\
+ \mathbb{E}_{B \sim p_{\text{data}}(B)} \left[ \| G_A(G_B(B)) - B \|_1 \right].
\end{multline}
The first term maintains the consistency in the Hazy $\to$ Clear $\to$ Hazy transformation, while the second term enforces consistency in the Clear $\to$ Hazy $\to$ Clear transformation.

\subsubsection{Cyclic Perceptual Consistency Loss}
To address the issue of perceptual quality and recover structural details that might be lost in pixel-wise losses, we introduce a Cyclic Perceptual Consistency loss. This is based on feature maps extracted from the 2nd and 5th pooling layers of the pre-trained VGG16 network \cite{vgg}. The perceptual loss is calculated using the Mean Squared Error (MSE) between the features of the original and generated images:
\begin{align}
\mathcal{L}_{\text{perceptual}} &= \| \phi_2(A) - \phi_2(G_B(G_A(A))) \|_2^2 \nonumber \\
&\quad + \| \phi_2(B) - \phi_2(G_A(G_B(B))) \|_2^2 \nonumber \\
&\quad + \| \phi_5(A) - \phi_5(G_B(G_A(A))) \|_2^2 \nonumber \\
&\quad + \| \phi_5(B) - \phi_5(G_A(G_B(B))) \|_2^2,
\end{align}
where \( \phi_2 \) and \( \phi_5 \) denote the feature extractors at the 2nd and 5th pooling layers of the VGG16 network, respectively.

\subsubsection{Cycle-Contrastive Loss}
To ensure content consistency by aligning generated outputs with input features while maximizing mutual information we introduce a Cycle-contrastive loss \( L_{\text{CCL}} \) for the training following (\ref{lccl}).

\subsection{Total Loss}
The total loss for the proposed model is a weighted ($\lambda$) sum of all the four components described so far.
\begin{align}
\mathcal{L}_{\text{total}} = & \lambda_{\text{GAN}} \mathcal{L}_{\text{GAN}} + \lambda_{\text{cyc}} \mathcal{L}_{\text{cyc}} + \lambda_{\text{perceptual}} \mathcal{L}_{\text{perceptual}} \nonumber \\ 
& + \lambda_{\text{contrastive}} \mathcal{L}_{\text{contrastive}},
\end{align}
where \(\lambda_{\text{GAN}}\) controls the weight of the adversarial loss, \(\lambda_{\text{cyc}}\) controls the weight of the cycle consistency loss,\(\lambda_{\text{perceptual}}\) controls the weight of the cyclic perceptual consistency loss, and \(\lambda_{\text{contrastive}}\) controls the weight of the contrastive loss.

For our experiments, we set the parameters as follows: \(\lambda_{\text{GAN}} = 2\), \(\lambda_{\text{cyc}} = 10\), and \(\lambda_{\text{perceptual}} = 0.1\), similar to \cite{aglc_gan}, while \(\lambda_{\text{contrastive}} = 0.3\) set experimentally to maintain generalizability.

\subsection{Elastic Weight Consolidation (EWC)}
\label{subsec:ewc}
Elastic Weight Consolidation (EWC) introduced in \cite{EWC} is a regularization-based continual learning method designed to mitigate catastrophic forgetting. EWC prevents catastrophic forgetting by selectively regularizing the network’s parameters based on their importance to earlier tasks. The importance of each parameter is measured using the Fisher Information Matrix (FIM), which quantifies how sensitive the model’s performance is to changes in specific parameters. Parameters deemed crucial for prior tasks are penalized more heavily, ensuring their preservation during the learning of new tasks.

The modified loss function in EWC is formulated as:
\begin{equation}
\mathcal{L}_{\text{EWC}} = \mathcal{L}_{\text{new}} + \frac{\lambda}{2} \sum_i F_i (\theta_i - \theta_i^*)^2,
\end{equation}
where \( \mathcal{L}_{\text{EWC}} \) is the Total loss function combining new task loss and regularization, \( \mathcal{L}_{\text{new}} \) the Loss for the current task, \( \lambda \) is the Regularization strength controlling the trade-off between retaining old task knowledge and learning the new task, \( F_i \) is the Fisher Information Matrix for parameter \( i \), indicating its importance in previous tasks, \( \theta_i \) is the Current value of parameter \( i \) during training, and \( \theta_i^* \) the Optimal value of parameter \( i \) from the previous task.

Elastic Weight Consolidation operates by first training the network on an initial task (e.g., DeHazing), after which the Fisher Information Matrix is computed to determine the importance of each parameter. These importance values are then stored and used to construct a regularization term in the loss function for subsequent tasks. During training on new tasks (e.g., DeSnowing and DeRaining), this regularization term penalizes deviations in critical parameters, ensuring knowledge retention while still allowing adaptation to new tasks. By integrating EWC into our framework, we ensure that our model effectively generalizes across different adverse weather conditions while preventing catastrophic forgetting.

\section{Dataset and Implementation Details}
\label{sec:implementation_details}
We first outline the datasets used in the experiments, followed by a discussion on the implementation details.

\subsection{Datasets Used}
\label{subsec:datasets_used}
\subsubsection{RESIDE – Dehazing Dataset \cite{li2019benchmarking}}
The Realistic Single Image Dehazing (RESIDE) dataset serves as a comprehensive benchmark for evaluating dehazing algorithms. It provides a diverse collection of synthetic and real-world hazy images. The dataset is divided into multiple subsets to evaluate the proposed model:

ITS (Indoor Training Set) Comprises of 13,990 synthetic hazy images and 1,399 corresponding clear images. This subset is created using a physical haze model that combines clean images with different haze densities. The controlled nature of the dataset makes it suitable for supervised training of dehazing networks.

SOTS (Synthetic Objective Testing Set) consists of 500 images, offering a balanced test set to evaluate the generalization capability of dehazing models on synthetic hazy images.

\subsubsection{SRRS – Desnowing Dataset \cite{JSTASRChen}}
The Snow Removal in Realistic Scenarios (SRRS) dataset is used to evaluate desnowing algorithms. The SRRS consists of real-world images captured in diverse snowy environments. The dataset includes 2,500 training images and 2,000 test images, covering various snow intensities and accumulation patterns.

\subsubsection{Rain100H – Deraining Dataset \cite{Yang2017RainRemoval}}
The Rain100H dataset is a benchmark for evaluating deraining models, containing rain-streaked images of varying intensities and orientations. Rain100H (Heavy Rain)contains 100 test images presenting challenging scenarios, where dense and overlapping rain streaks obscure objects and textures in the scene, requiring more advanced deraining techniques.

\subsection{Implementation Details}
We implement the proposed DA-AGLC-GAN model on the PyTorch framework and trained on NVIDIA RTX A4000. Training is conducted for 60,000 iterations on all the datasets mentioned in Section \ref{subsec:datasets_used}. The Adam optimizer (momentum = 0.5) is used with an initial learning rate of 0.0001, which gradually decays to 0.00005 following a cosine decay schedule, with a batch size of 1 maintained throughout training.

Data augmentation is applied, where images are randomly cropped and scaled to a resolution of $256 \times 256$ before being fed into the model. Since CycleGAN forms the backbone of the architecture, the order in which images are fed does not impact training. For efficiency, both hazy and corresponding clear images are simultaneously input into separate branches of the model, allowing for a more streamlined and structured training approach.

The performance of the model is assessed using two widely accepted metrics: \textbf{PSNR} (Peak Signal-to-Noise Ratio) and \textbf{SSIM} (Structural Similarity Index).
In our experiments, \(R\), is implicitly set to \(1.0\) because the images are normalized to the \([0, 1]\) range.
\begin{table}
\centering
\caption{Comparisons of SOTA Models over RESIDE ITS.}
\label{tab:reside_ITS}
\renewcommand{\arraystretch}{1.3} 
\setlength{\tabcolsep}{12pt} 
\begin{tabular}{lcc}
\hline
\textbf{Model} & \textbf{PSNR} & \textbf{SSIM} \\
\hline
\multicolumn{3}{c}{\textbf{Paired Models}} \\
AOD-NET \cite{AOD-NET}         & 19.06 & 0.8504 \\
DehazeNet \cite{DeHazeNet}       & 21.14 & 0.8472 \\
FFA-NET \cite{ref_ffa_net}          & 36.39 & 0.9886 \\
ACER-NET \cite{acerNet}        & 37.17 & 0.9901 \\
Su et al. \cite{SU2023103706}       & 31.77 & 0.9522 \\
CL2S \cite{rohn2024rethinkingelementaryfunctionfusion} & 35.36  & 0.9808\\
Omni Kernal Network \cite{Cui_Ren_Knoll_2024}  & 40.79 &  0.996\\
CasDyF-Net \cite{yinglong2024casdyfnetimagedehazingcascaded} &  43.21 & 0.997 \\
\hline
\multicolumn{3}{c}{\textbf{Unpaired Models}} \\
DCP \cite{ffa_net}             & 16.62 & 0.8179 \\
Improved CycleGAN \cite{CHAITANYA2021103014} & 20.05 & 0.8307 \\
Yang et al. \cite{Yang_2022_CVPR}      & 25.42 & 0.9320 \\
Chen et al. \cite{chen}      & 24.61 & 0.9180 \\
Ren et al. \cite{Ren2022-fg} & 29.04 &  0.8705\\
Fan et al. \cite{FAN2025150} & 26.23 & 0.932 \\
Jaisurya et al. \cite{jaiSurya_0}  & 31.67 & 0.9612 \\
AGLC-GAN. \cite{aglc_gan}   & 31.69 & 0.9639 \\
\hline
\textbf{DA-AGLC-GAN} (ours) & \textbf{32.31} & \textbf{0.9697} \\
\hline
\end{tabular}
\end{table}

\begin{table}
\centering
\caption{Comparisons of SOTA Models over RESIDE OTS.}
\label{tab:reside_OTS}
\renewcommand{\arraystretch}{1.3}
\setlength{\tabcolsep}{12pt}
\begin{tabular}{lcc}
\hline
\textbf{Model} & \textbf{PSNR} & \textbf{SSIM} \\
\hline
\multicolumn{3}{c}{\textbf{Paired Models}} \\
AOD-NET \cite{AOD-NET}       & 20.29 & 0.8765 \\
DehazeNet \cite{DeHazeNet}     & 22.46 & 0.8514 \\
FFA-NET \cite{ref_ffa_net}        & 33.57 & 0.9840 \\
Su et al. \cite{SU2023103706}      & 33.41 & 0.9698 \\
Omni Kernal Network \cite{Cui_Ren_Knoll_2024}  & 37.68 & 0.995\\
CasDyF-Net \cite{yinglong2024casdyfnetimagedehazingcascaded} &  38.86 & 0.995 \\
MixDehazeNet \cite{lu2023mixdehazenetmixstructure} & 30.18 & 0.973\\

\hline
\multicolumn{3}{c}{\textbf{Unpaired Models}} \\
DCP \cite{ffa_net}           & 19.13 & 0.8148 \\
Improved CycleGAN \cite{CHAITANYA2021103014}& 21.14 & 0.8919 \\
UCL-DeHaze \cite{UCL-Dehaze} & 25.21 &0.927 \\
Yang et al. \cite{Yang_2022_CVPR}    & 25.83 & 0.9560 \\
Chen et al. \cite{chen}    & 24.61 & 0.9180 \\
Dehaze-GLCGAdN \cite{deHaze-glcgan}   & 26.51 & 0.9354 \\
Fan et al. \cite{FAN2025150} & 26.52 & 0.951 \\
Jaisurya et al. \cite{jaiSurya_0} & 36.17 & 0.9745 \\
AGLC-GAN. \cite{aglc_gan}   & 36.71 & 0.9781 \\
\hline
\textbf{DA-AGLC-GAN} (ours) & \textbf{37.13} & \textbf{0.9793} \\
\hline
\end{tabular}
\end{table}

\section{Results and Discussions}
\label{sec:results_discussions}
We evaluate the model on a haze dataset, a snow dataset, and a rain dataset to assess its effectiveness on each restoration task independently. Next, we describe our experiments on the continual learning setup aimed at generalizing the model across deHaze, deSnow, and deRain tasks. Finally, we present ablation studies for a better analysis of the proposed method.
\begin{figure*}
    \centering
    \includegraphics[width=\textwidth]{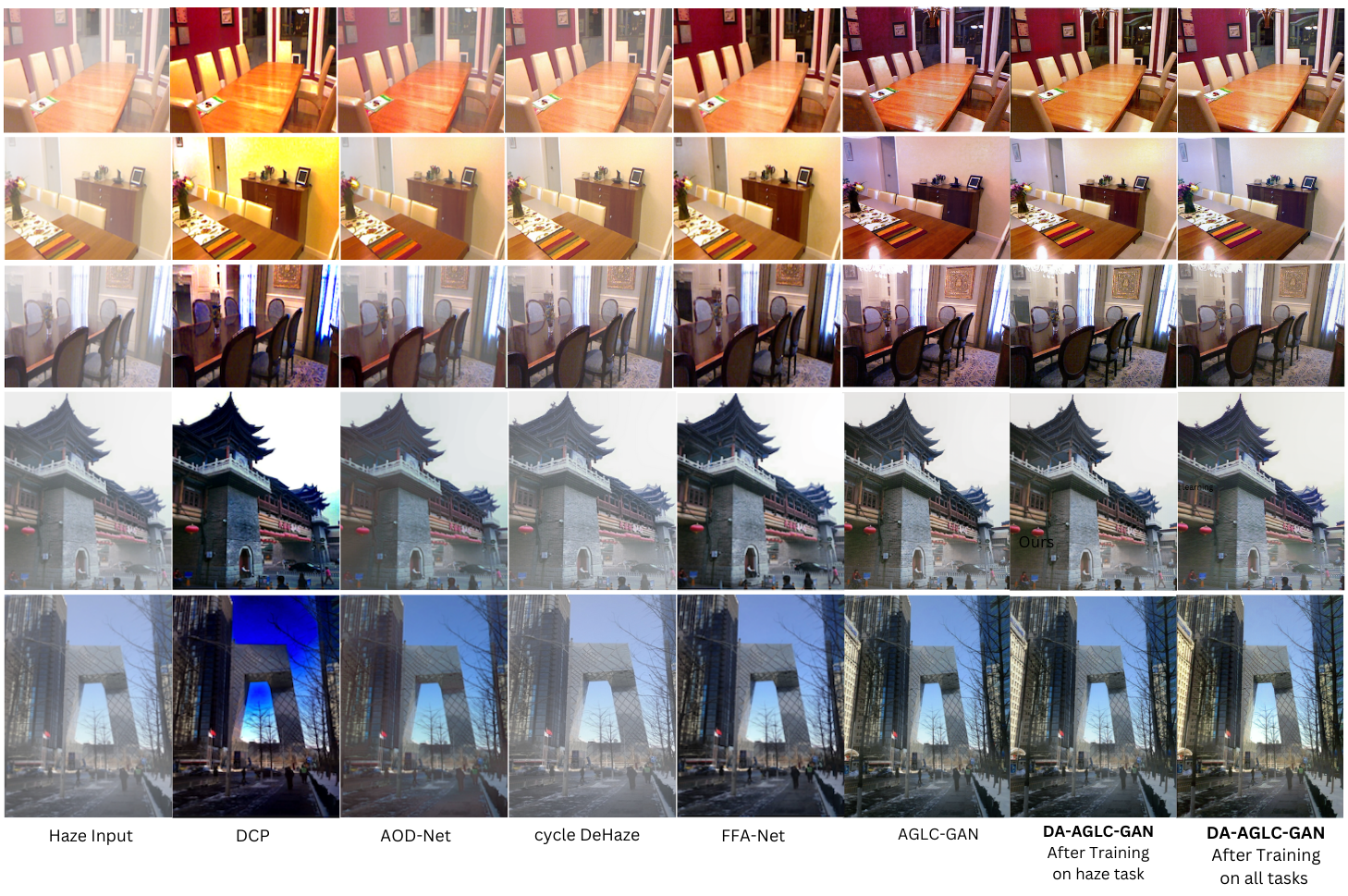}
    \caption{Qualitative comparisons of the proposed DA-AGLC-GAN with state-of-the-art (SOTA) models. The first three rows illustrate results on indoor images, while the last two rows depict results on outdoor images.}
    \label{fig:comparision_sota}
\end{figure*}

\subsection{Results on RESIDE Dataset (DeHazing)}
\label{subsec:reside_results}
The experiments on RESIDE dataset \cite{li2019benchmarking} were conducted on two splits: RESIDE ITS and RESIDE OTS. For RESIDE ITS, our model was trained for 5 epochs with 13k samples per epoch, and for RESIDE OTS, training was performed for 1 epoch with 60k samples per epoch. Tables \ref{tab:reside_ITS} and \ref{tab:reside_OTS} present quantitative comparisons for both ITS and OTS Datasets respectively in terms of PSNR and SSIM, where the first set of models are supervised (paired) methods and the latter are unsupervised (unpaired) approaches. The proposed model performs better compared to the competing unpaired methods, for both ITS and OTS datasets.

The qualitative results of the proposed model are shown in Fig. \ref{fig:comparision_sota} when applied to the ITS and OTS datasets, respectively. We can observe that the proposed model, due to its attention mechanism and updated fusion module, provides closer results to the ground truth while preserving the original color and texture information of the images.
\begin{table}
\centering
\caption{Comparisons of SOTA Models over SRRS Dataset.}
\label{tab:results_desnow_sota}
\renewcommand{\arraystretch}{1.3}
\setlength{\tabcolsep}{12pt}
\begin{tabular}{lcc}
\hline
\textbf{Model} & \textbf{PSNR} & \textbf{SSIM} \\
\hline
\multicolumn{3}{c}{\textbf{Paired Models}} \\
DesnowNet \cite{DesnowNet} & 21.30 & 0.835\\
SMGARN \cite{cheng2022snowmaskguidedadaptive} &  25.43 & 0.92\\
LMQFormer \cite{LMQFormer}  &  31.04 & 0.964 \\
Conv-IR \cite{Conv-IR} & 32.39 & 0.98 \\
\hline
\multicolumn{3}{c}{\textbf{Unpaired Models}} \\
Cycle-GAN \cite{CycleGAN_deHAze} & 20.21 & 0.74 \\
AGLC-GAN \cite{aglc_gan}   & 34.65 & 0.9357 \\
\hline
\textbf{DA-AGLC-GAN} (ours) & \textbf{35.53} & \textbf{0.9432} \\
\hline
\end{tabular}
\end{table}
\begin{figure}
    \centering
    \includegraphics[width=0.45\textwidth]{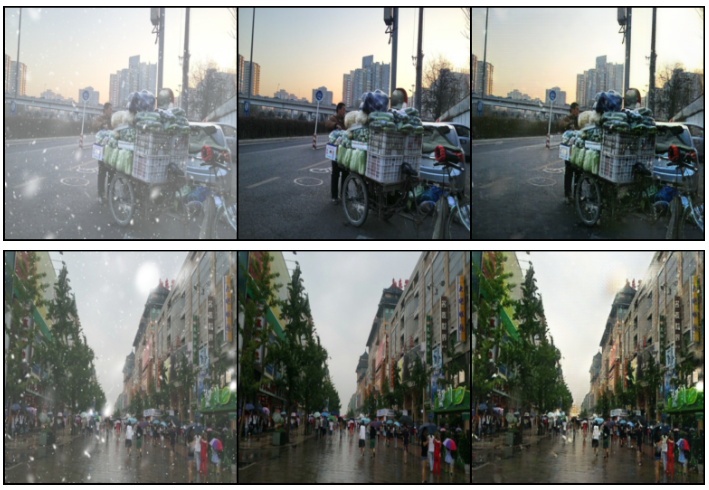}
\caption{ Results of the proposed model on SRRS dataset. (a) Input Snow image. (b) Ground Truth Clear Image. (c) Generated Clear Image.}
    \label{fig:snow_image}
\end{figure}

\subsection{Results on SRRS Dataset (DeSnowing)}
\label{subsec:srrs_results}
The proposed approach successfully restores images degraded by snow, as evidenced by the improvements in both PSNR and SSIM scores on SRRS dataset \cite{JSTASRChen} (see Table~\ref{tab:results_desnow_sota}). In particular, the proposed DA-AGLC-GAN performs even better than the paired models. These results highlight the effectiveness of our unsupervised approach in striking a balance between high-fidelity restoration and structural consistency. The qualitative results of the proposed model are shown in Fig. \ref{fig:snow_image} when applied to the SRRS dataset.
\begin{table}
\centering
\caption{Comparisons of SOTA Models over Rain100H Dataset.}
\label{tab:results_derain_sota}
\renewcommand{\arraystretch}{1.3}
\setlength{\tabcolsep}{12pt}
\begin{tabular}{lcc}
\hline
\textbf{Model} & \textbf{PSNR} & \textbf{SSIM} \\
\hline
\multicolumn{3}{c}{\textbf{Paired Models}} \\
RESCAN \cite{RE-SCAN} & 26.45 & 0.8458\\
MCW-Net \cite{mcw-net} & 30.70 & 0.922	\\
PReNet \cite{preNet} & 29.46 & 0.899\\
Restormer \cite{restormer} & 31.46 & 0.904\\
IR-SDE \cite{luo2023imagerestorationmeanrevertingstochastic} & 31.65 & 0.9041 \\
DA-CLIP \cite{luo2024controllingvisionlanguagemodelsmultitask} & 33.91 & 0.926\\
\hline
\multicolumn{3}{c}{\textbf{Unpaired Models}} \\
 wei wei et al. \cite{8954172}  & 16.56 & 0.486	 \\
AGLC-GAN \cite{aglc_gan}   & 31.79 & 0.8310 \\
\hline
\textbf{DA-AGLC-GAN} (ours) & \textbf{32.23} & \textbf{0.8434} \\
\hline
\end{tabular}
\end{table}
\begin{figure}
    \centering
    \includegraphics[width=0.45\textwidth]{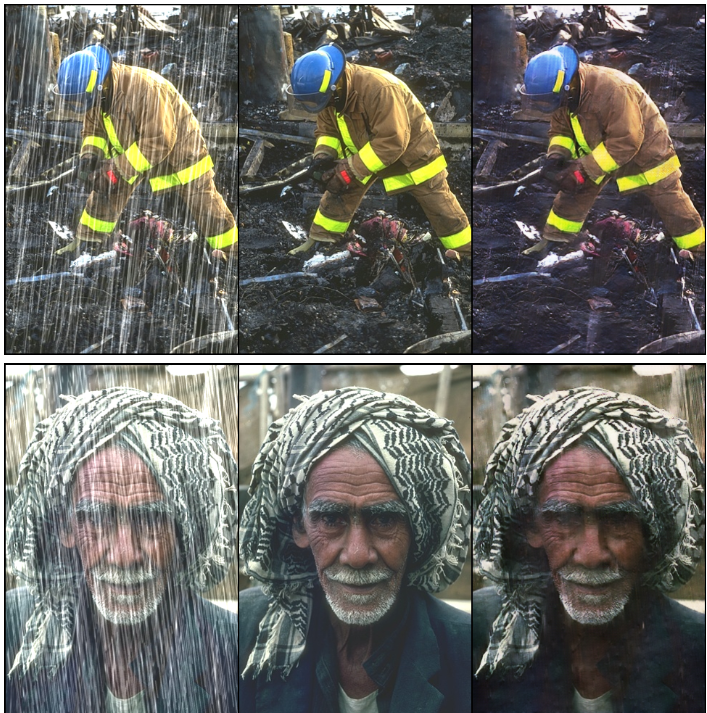}
\caption{ Results of the proposed model on ITS/OTS dataset. (a) Input rainy image. (b) Ground Truth Clear Image. (c) Generated Clear Image.}
    \label{fig:rain_image}
\end{figure}
\begin{figure*}
    \centering
    \includegraphics[width=0.75\textwidth]{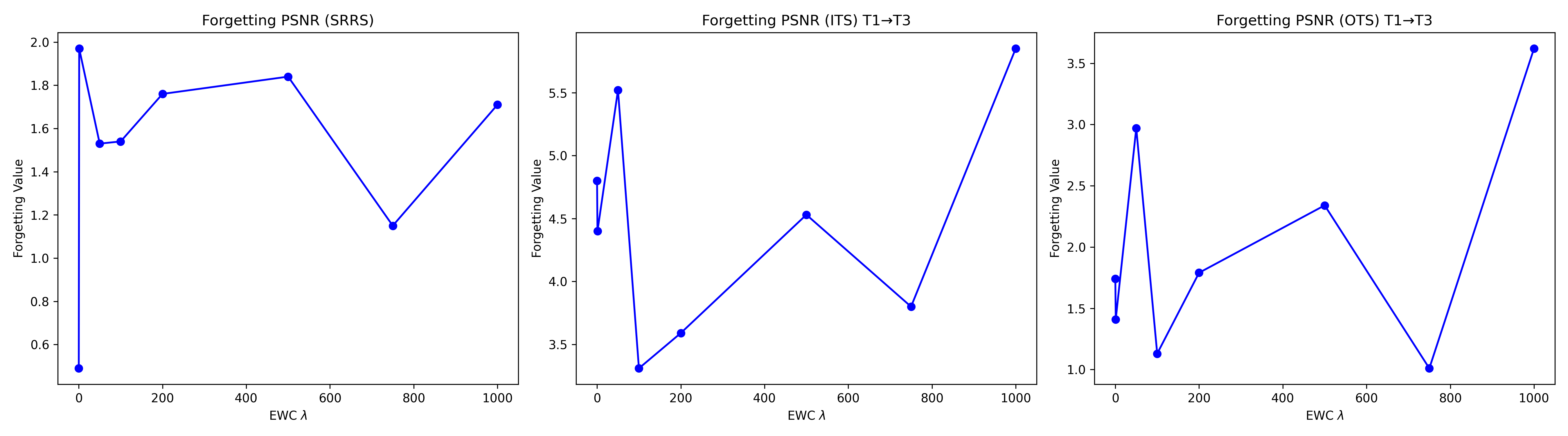}
    \caption{Variation of forgetting with EWC \(\lambda\). Lower forgetting values indicate better retention of previously learned tasks.}
    \label{fig:ewc_forgetting}
\end{figure*}
\begin{figure}
    \centering
    \includegraphics[width=0.75\columnwidth]{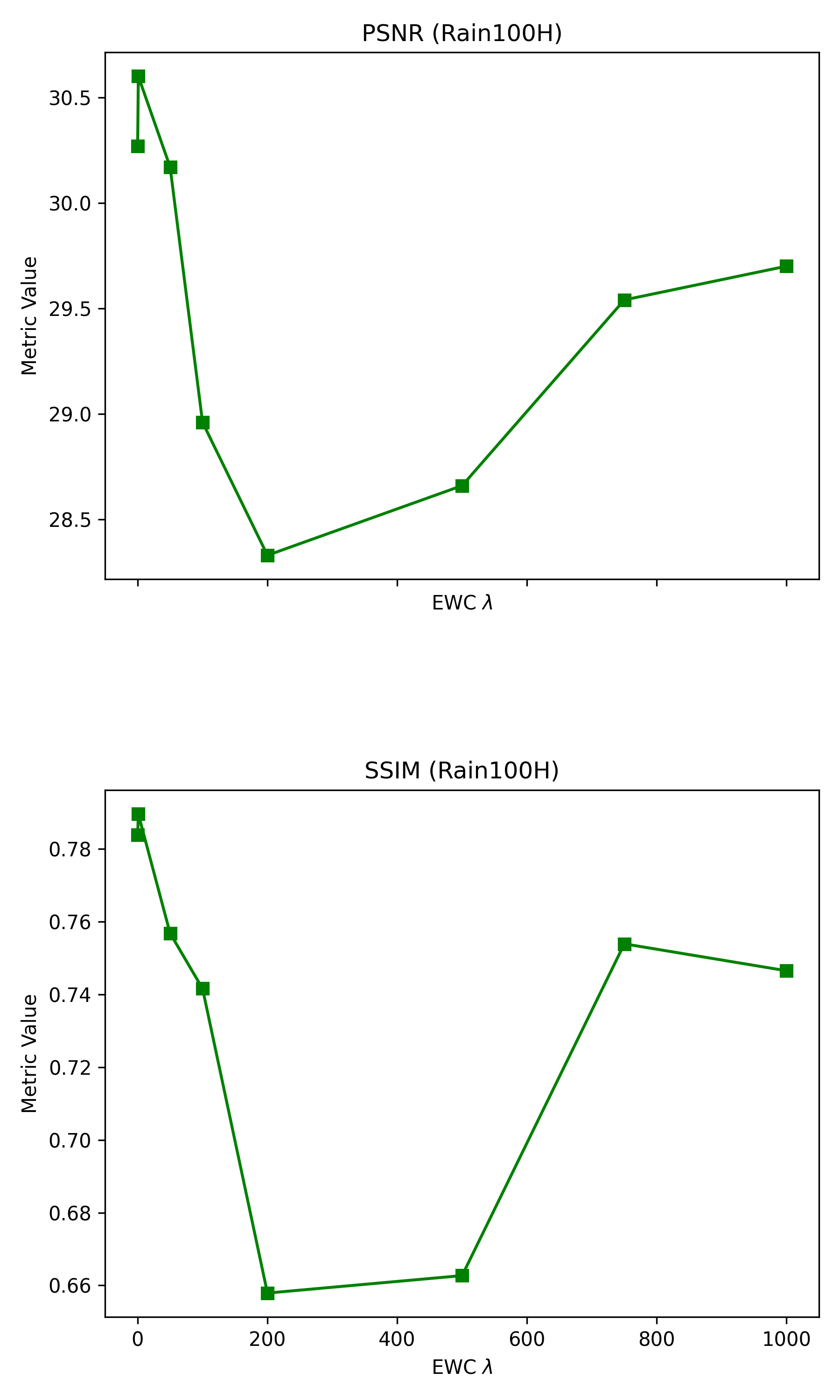}
    \caption{Final PSNR performance on deSnow (Task 2) and deRain (Task 3) as a function of EWC \(\lambda\).}
    \label{fig:ewc_performance}
\end{figure}

\subsection{Rain 100H (DeRaining)}
\label{subsec:rain100H_results}
The performance of the proposed method achieves superior results compared to existing methods for Rain 100H dataset \cite{Yang2017RainRemoval}, confirming its robustness in handling rain-induced degradations (refer to Table \ref{tab:results_derain_sota}). The proposed method outperforms all the unpaired models for deraining. The qualitative results of the proposed model are shown in Fig. \ref{fig:rain_image} when applied to the Rain100H dataset.
\begin{table*}
\centering
\scriptsize 
\setlength{\tabcolsep}{4pt} 
\renewcommand{\arraystretch}{1.2} 
\caption{Comparison of EWC Configurations for Task 3 (deRain) with Detailed Forgetting Metrics.}
\label{tab:ewc_config_comparison}
\begin{tabular}{lccccccc}
\hline
\textbf{Configuration} 
& \textbf{F-PSNR (ITS T1$\rightarrow$2)} 
& \textbf{F-SSIM (ITS T1$\rightarrow$2)} 
& \textbf{F-PSNR (SRRS)} 
& \textbf{F-SSIM (SRRS)} 
& \textbf{F-PSNR (ITS T1$\rightarrow$3)} 
& \textbf{F-SSIM (ITS T1$\rightarrow$3)} \\
\hline
$\lambda_1 = 700, \lambda_2 = 800$ 
& 3.39 
& 0.0131 
& 1.49 
& 0.0133 
& 4.96 
& 0.0404 \\

$\lambda_1 = 800, \lambda_2 = 700$ 
& 3.55 
& 0.0121 
& 1.53 
& 0.0143 
& 5.15 
& 0.0411 \\

\hline
\end{tabular}
\end{table*}

\subsection{Parameter Analysis: EWC $\lambda$}
\label{subsec:ewc_lambda}
To explore the impact of $\lambda$ on the proposed EWC, we vary \(\lambda\) from \(0\) to \(1000\) to find the suitable value. After training on each task, we evaluated the final performance (PSNR and SSIM) and measured the degree of \emph{forgetting} for previously learned tasks.
\begin{table}
\centering
\caption{Dehazing Results on RESIDE ITS and OTS Datasets}
\label{tab:results_dehaze}
\begin{tabular}{lcccc}
\hline
\textbf{Method} & \textbf{Train} & \textbf{Test} & \textbf{PSNR} & \textbf{SSIM} \\
\hline
\textbf{AGLC-GAN} & ITS & ITS test & 31.69 & 0.9639 \\
                  & ITS & OTS test & 33.50 & 0.9521 \\
                  & OTS & OTS test & 36.71 & 0.9781 \\
                  & OTS & ITS test & 27.91 & 0.9588 \\
\hline
\textbf{AGLC-GAN + SK-Fusion} & ITS & ITS test & 31.76 & 0.9654 \\
                             & ITS & OTS test & 33.43 & 0.9605 \\
                             & OTS & OTS test & 36.30 & 0.9776 \\
                             & OTS & ITS test & 28.09 & 0.9571 \\
\hline
\textbf{AGLC-GAN + CL-Loss} & ITS & ITS test & 30.74 & 0.9634 \\
                           & ITS & OTS test & 34.27 & 0.9571 \\
                           & OTS & OTS test & 36.91 & 0.9789 \\
                           & OTS & ITS test & 28.21 & 0.9582 \\
\hline
\textbf{DA-AGLC-GAN} & ITS & ITS test & \textbf{32.31} & \textbf{0.9697} \\
                    & ITS & OTS test & \textbf{35.08} &\textbf{ 0.9660} \\
                    & OTS & OTS test & \textbf{37.13} & \textbf{0.9793} \\
                    & OTS & ITS test & \textbf{28.25} & \textbf{0.9593} \\
\hline
\end{tabular}
\end{table}

\begin{table}
\centering
\caption{DeSnowing Results on the SRRS Dataset}
\label{tab:results_desnow}
\begin{tabular}{lcccc}
\hline
\textbf{Method} & \textbf{Train Dataset} &  \textbf{PSNR} & \textbf{SSIM} \\
\hline
AGLC-GAN & SRRS & 34.65 & 0.9357 \\
AGLC-GAN + SK-Fusion & SRRS & 34.96 & \textbf{0.9437} \\
AGLC-GAN + CL-Loss & SRRS & 34.69 & 0.9309 \\
DA-AGLC-GAN & SRRS  & \textbf{35.53} & 0.9432 \\
\hline
\end{tabular}
\end{table}

\begin{table}
\centering
\caption{DeRaining Results on the Rain 100H Dataset}
\label{tab:results_derain}
\begin{tabular}{lcccc}
\hline
\textbf{Method} & \textbf{Train Dataset} &  \textbf{PSNR} & \textbf{SSIM} \\
\hline
AGLC-GAN & Rain 100H & 31.79 & 0.8310 \\
AGLC-GAN + SK-Fusion & Rain 100H & 32.37 & 0.8387 \\
AGLC-GAN + CL-Loss & Rain 100H & 32.18 & 0.8416 \\
DA-AGLC-GAN & Rain 100H &  \textbf{32.33} & \textbf{0.8434} \\
\hline
\end{tabular}
\end{table}

\paragraph{Defining Forgetting.}
If \(P_{T_1}^{\text{end}}\) denotes the performance at the end of Task 1 and \(P_{T_1}^{\text{postT3}}\) represents the performance on Task 1 after training on Tasks 2 and 3, then:
\begin{equation}
Forgetting = P_{T_1}^{end} - P_{T_1}^{postT3}.
\end{equation}
A lower or negative value suggests better retention of the previous knowledge.
\begin{figure*}
    \centering
    \includegraphics[width=\textwidth]{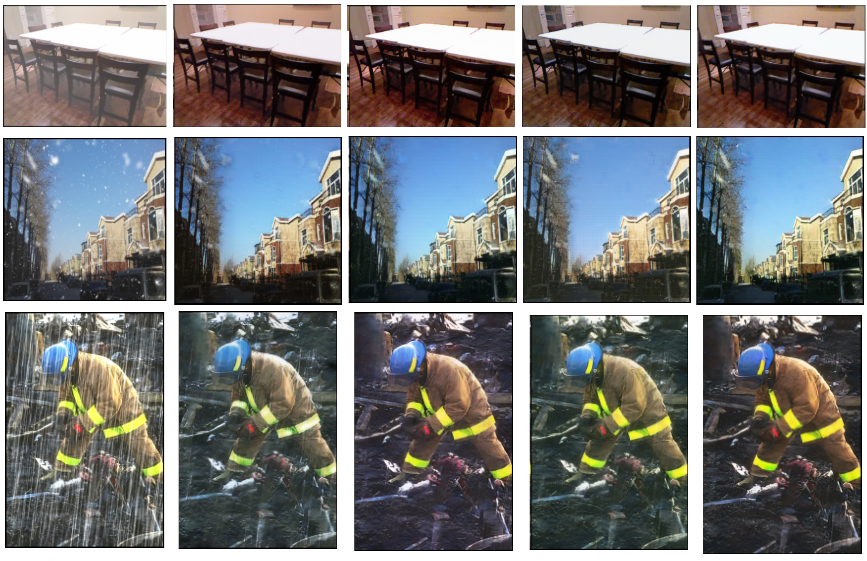}
\caption{ Results of the proposed model on all datasets. (a) Input image. (b) Output of AGLC-GAN. (c) Output of AGLC-GAN + SkFusion (d) Output of AGLC-GAN + ccl loss (e) Output of DA-AGLC-GAN.}
    \label{fig:comparison}
\end{figure*}

\paragraph{Observations and Plots.}
Figure~\ref{fig:ewc_forgetting} shows the variation of forgetting with different \(\lambda\) values, and Figure~\ref{fig:ewc_performance} illustrates the final PSNR and SSIM performance on Task 3 (deRain) as a function of \(\lambda\). Our findings reveal that very small \(\lambda\) values (e.g., 0.5 or 1) do not adequately mitigate catastrophic forgetting, whereas extremely high values (e.g., 1000) over-constrain parameter updates and hamper adaptation to new tasks. Moderate \(\lambda\) values provide an optimal balance between retaining previously learned knowledge and facilitating new learning. Based on our experiments, the optimal \(\lambda\) for our model is 750.

Given this optimal value, we further investigate the impact of the relative settings of \(\lambda_1\) and \(\lambda_2\), where \(\lambda_1\) is applied after training Task 1 (deHaze) and \(\lambda_2\) after training Task 2 (deSnow). Table~\ref{tab:ewc_config_comparison} compares two configurations: \(\lambda_1=700, \lambda_2=800\) versus \(\lambda_1=800, \lambda_2=700\). Our results indicate that setting \(\lambda_2\) greater than \(\lambda_1\) yields slightly better performance in terms of both PSNR and SSIM on the deRain task, while also exhibiting reduced forgetting of Task 1.

In summary, our results demonstrate that careful tuning of the EWC \(\lambda\) parameter is critical for balancing knowledge retention and the ability to learn new tasks. Specifically, moderate \(\lambda\) values—especially with \(\lambda_2\) set higher than \(\lambda_1\)—yield the best performance on the deRain task alongside reduced forgetting. These findings underscore the effectiveness of our continual learning framework in achieving robust and generalized performance across diverse adverse weather conditions. The qualitative comparison of dehazing performance for our model, trained solely on haze data as well as after training on all three tasks using continual methods, is shown in Figure~\ref{fig:comparision_sota}.

\subsection{Ablation Studies}
\label{subsec:ablation_studies}
We first trained and tested our model on different tasks separately to evaluate its performance in adverse weather removal. Table~\ref{tab:results_dehaze} presents the dehazing results on the RESIDE ITS and RESIDE OTS datasets, while Table~\ref{tab:results_desnow} and Table~\ref{tab:results_derain} show the deSnowing and deRaining results on the SRRS and Rain 100H datasets, respectively. Our proposed DA-AGLC-GAN consistently outperforms baseline methods in terms of PSNR and SSIM. The Qualitative comparisons of Haze, Snow and Rain Removal has been shown in Fig ~\ref{fig:comparison}. We can observe the efficacy of DA-AGLC-GAN over the base model AGLC-GAN.

\section{Conclusion and Future Work}
\label{sec:conclusion}
We presented All in one-  DA-AGLC-GAN, a unified framework for adverse weather removal that significantly improves upon the baseline AGLC-GAN by integrating selective kernel fusion layers, cycle-contrastive learning, and Elastic Weight Consolidation (EWC) for continual learning. Our approach not only enhances restoration quality across deHaze, deSnow, and deRain tasks but also effectively mitigates catastrophic forgetting, ensuring stable performance when sequentially learning from multiple adverse weather conditions. Future work may investigate transformer-based architectures and self-supervised learning strategies to further boost the generalization and efficiency of the model. In addition, exploring adaptive mechanisms to balance stability and plasticity during continual learning could yield further performance gains, ultimately enabling more robust image restoration in dynamic, real-world environments.

\bibliographystyle{IEEEtran}
\bibliography{references}

\begin{IEEEbiography}
[{\includegraphics[width=1in,height=1.25in,clip,keepaspectratio]{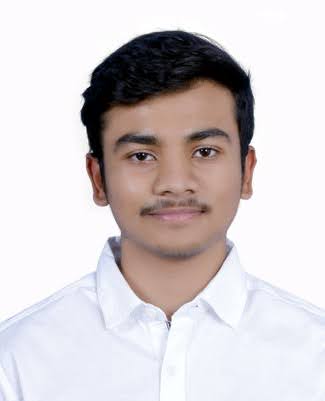}}]{Kotha Kartheek}{\space} is a senior at the Shiv Nadar Institute of Eminence in Delhi NCR, India, where he is pursuing a Bachelor of Technology in Computer Science and Engineering, along with a minor in Mathematics. He is currently interning as a Software Engineer at Chegg Inc., contributing to cutting-edge software development projects. Kartheek has built a strong expertise in computer vision through projects like Action Recognition and Adverse Weather Removal. He is also passionate about exploring how machine learning can be applied across different fields; for example, he is working on a project that uses ML techniques to improve handover reliability in communication networks. After completing his undergraduate studies, Kartheek plans to pursue graduate studies to further deepen his understanding of machine learning and to work toward a research career. 

\end{IEEEbiography}

\begin{IEEEbiography}
[{\includegraphics[width=1in,height=1.25in,clip,keepaspectratio]{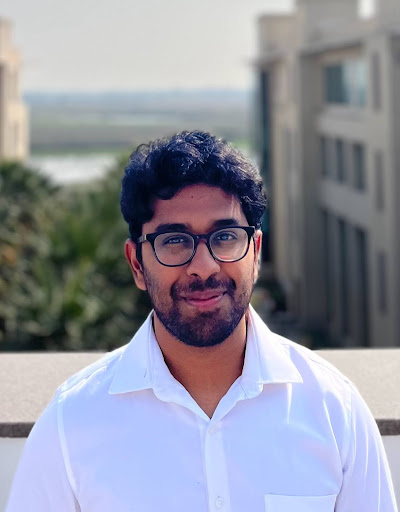}}]{Lingamaneni Gnanesh Chowdary}{\space} is pursuing a Bachelor of Technology in Computer Science and Engineering at Shiv Nadar Institute of Eminence, Delhi NCR, India. Passionate about Deep Learning, Gnanesh has actively engaged in innovative projects such as Adverse Weather Removal using Cycle-GANs and anomaly detection in vision tasks. His leadership and organizational skills have been demonstrated through his role as an Admission Coordinator at Shiv Nadar IoE, and his commitment to academic excellence is reflected in his recognition on the Dean’s List. Gnanesh’s academic interests span Computer Science fundamentals, Software Development, Machine Learning, and he is driven by a desire to explore and contribute to cutting-edge research in these fields. 

\end{IEEEbiography}

\begin{IEEEbiography}
[{\includegraphics[width=1in,height=1.25in,clip,keepaspectratio]{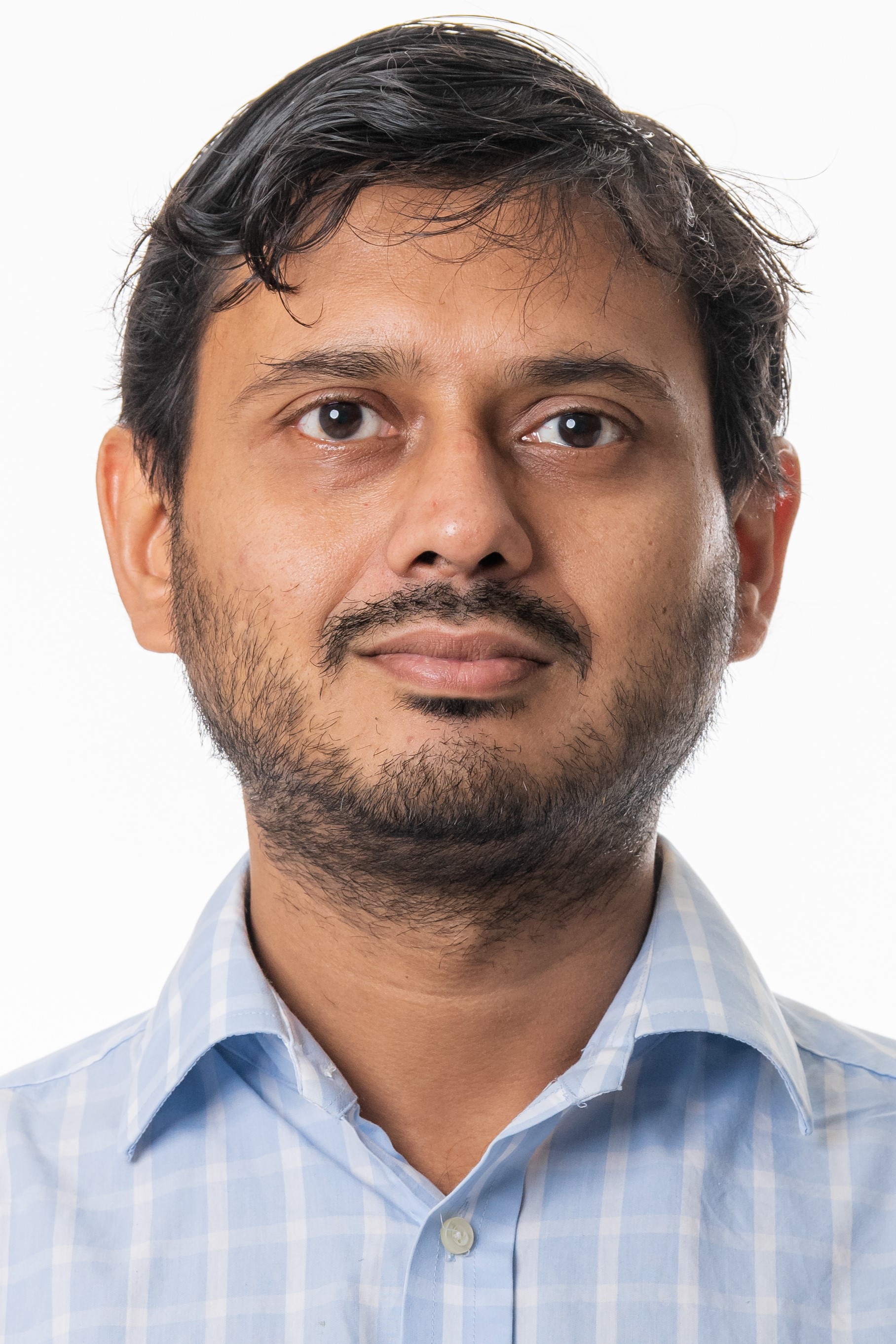}}]{Snehasis Mukherjee}{\space} Snehasis Mukherjee (Senior Member, IEEE) is an Associate Professor at Shiv Nadar Institution of Eminence, Delhi NCR, India. He obtained his PhD in Computer Science from Indian Statistical Institute in 2012. He worked as a Post Doctoral Fellow at NIST, Gaithersburg, USA. Then he spent 6 years as an Assistant Professor at IIIT Sri City till April 2020. He has authored several peer-reviewed research papers in reputed journals and conferences. He is an active reviewer of several reputed journals such as IEEE Trans. NNLS, IEEE Trans. CSVT, IEEE Trans. IP, IEEE Trans. ETCI, IEEE Trans. HMS, IEEE Trans. Cyb, IEEE CIM, CVIU, Pattern Recognition, Neural Networks, Neurocomputing, and many more, and conferences such as BMVC, WACV, etc. His research area includes Computer Vision, Machine Learning, Image Processing and Graphics.

\end{IEEEbiography}

\end{document}